%% file: main.tex
\documentclass[10pt,twocolumn,letterpaper]{article}

\usepackage[pagenumbers]{iccv} 
\usepackage{graphicx}
\usepackage{booktabs}
\usepackage{multirow}


\definecolor{iccvblue}{rgb}{0.21,0.49,0.74}
\usepackage[pagebackref,breaklinks,colorlinks,allcolors=iccvblue]{hyperref}

\def\paperID{9567} 
\def\confName{ICCV}
\def\confYear{2025}

\title{M-DocSum: Do LVLMs Genuinely Comprehend Interleaved Image-Text in Document Summarization?}

\author{
Haolong Yan\textsuperscript{1,2}\thanks{Core Contribution. Email: \texttt{yanhaolong@bupt.edu.cn}}\qquad
Kaijun Tan\textsuperscript{2}\footnotemark[1]\qquad
Yeqing Shen\textsuperscript{2}\qquad
Xin Huang\textsuperscript{2,3}\qquad
Zheng Ge\textsuperscript{2}\thanks{Corresponding Author}\\
Xiangyu Zhang\textsuperscript{2}\qquad
Si Li\textsuperscript{1}\footnotemark[2]\qquad
Daxin Jiang\textsuperscript{2}\\
\textsuperscript{1}BUPT \quad \textsuperscript{2}StepFun \quad \textsuperscript{3}Waseda University
}

\makeatletter
\newcommand{\customfootnote}[1]{%
    \begingroup
    \renewcommand{\thefootnote}{\relax}
    \renewcommand{\@makefntext}[1]{\noindent\hspace*{1em}\textsuperscript{#1} }
    \footnotetext{#1}%
    \endgroup
}
\makeatother

\begin{document}

\twocolumn[{
\renewcommand\twocolumn[1][]{#1}
\maketitle
\begin{center}
    \centering
    \captionsetup{type=figure}
    \includegraphics[width=\textwidth]{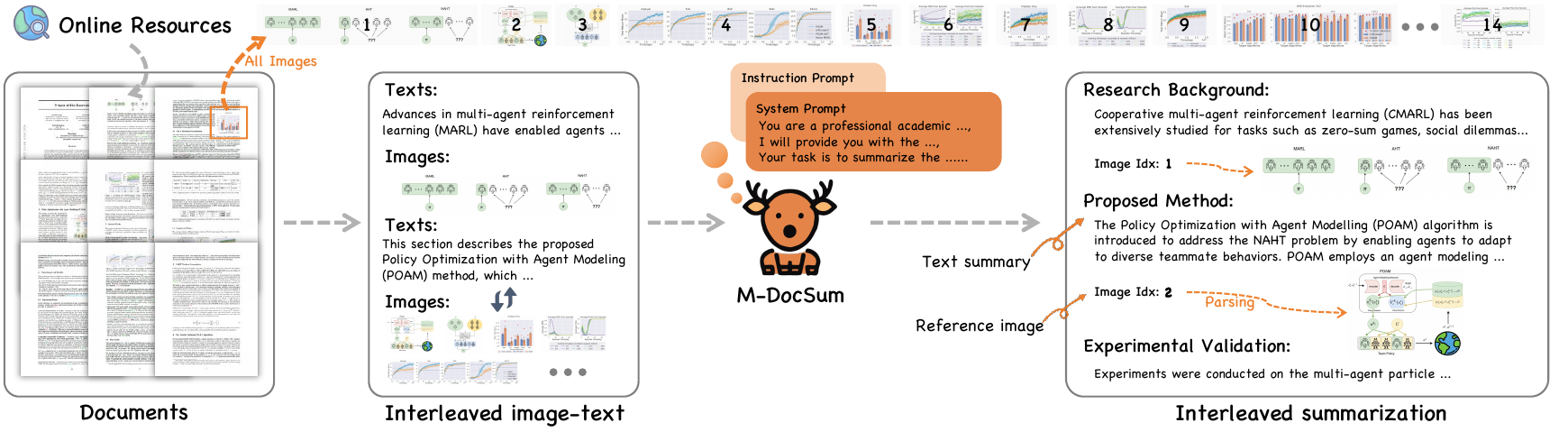}
    \captionof{figure}{Illustration of the Interleaved Multimodal Summarization task. The document is an example from M-DocSum-Bench, which has $24$ pages and a total of $14$ images. The Interleaved Summarization is generated by our M-DocSum-7B model.}
    \label{fig:intro}
\end{center}
}]

\customfootnote{*~Core Contribution. Email: \texttt{yanhaolong@bupt.edu.cn}}
\customfootnote{\textdagger~Corresponding Author.} 

\input{sec/0_abstract}    
\input{sec/1_intro}

\input{sec/2_related_work}
\input{sec/3_intersum_bench}
\input{sec/4_training}
\input{sec/5_experiments}
\input{sec/6_conclusion}

{
    \small
    \bibliographystyle{ieeenat_fullname}
    \bibliography{main}
}

\input{sec/7_supp}

\end{document}

%% file: sec/0_abstract.tex
\begin{abstract}
We investigate a critical yet under-explored question in Large Vision-Language Models~(LVLMs): Do LVLMs genuinely comprehend interleaved image-text in the document?
Existing document understanding benchmarks often assess LVLMs using question-answer formats, which are information-sparse and difficult to guarantee the coverage of long-range dependencies.
To address this issue, we introduce a novel and challenging Multimodal Document Summarization Benchmark~(M-DocSum-Bench), which comprises 500 high-quality arXiv papers, along with interleaved multimodal summaries aligned with human preferences.
M-DocSum-Bench is a reference-based generation task and necessitates the generation of interleaved image-text summaries using provided reference images, thereby simultaneously evaluating capabilities in understanding, reasoning, localization, and summarization within complex multimodal document scenarios.
To facilitate this benchmark, we develop an automated framework to construct summaries and propose a fine-grained evaluation method called M-DocEval.
Moreover, we further develop a robust summarization baseline, i.e., M-DocSum-7B, by progressive two-stage training with diverse instruction and preference data.
The extensive results on our M-DocSum-Bench reveal that the leading LVLMs struggle to maintain coherence and accurately integrate information within long and interleaved contexts, often exhibiting confusion between similar images and a lack of robustness. 
Notably, M-DocSum-7B achieves state-of-the-art performance compared to larger and closed-source models (including GPT-4o, Gemini Pro, Claude-3.5-Sonnet and Qwen2.5-VL-72B, etc.), demonstrating the potential of LVLMs for improved interleaved image-text understanding. 
The code, data, and models are available at \url{https://github.com/stepfun-ai/M-DocSum-Bench}.

\end{abstract}

%% file: sec/1_intro.tex
\section{Introduction}
The ability to generate concise, interleaved image-text summaries is crucial for a wide range of real-world applications~\cite{visualwebbench_s29, docvqa_s13}. 
Imagine automatically generating accessible presentations from complex scientific papers, producing interactive reports from data-heavy analyses, or even crafting engaging social media posts summarizing lengthy articles. 
These scenarios demand a deep understanding of not only the textual content but also the visual information and the intricate relationships between them. 
However, despite the importance of this capability, the field of interleaved image-text document summarization remains largely unexplored~\cite{chartqa_s15, towards_s28, infographicvqa_s17}. 
There is a critical lack of suitable benchmarks to evaluate the performance of Large Vision-Language Models (LVLMs) in this challenging setting, and consequently, a dearth of effective methodologies tailored for this task.
Therefore, a crucial question remains: \textbf{Do LVLMs genuinely comprehend interleaved image-text in documents? }
Addressing this question is vital for advancing document understanding towards more complex and practical applications.

Existing benchmarks primarily feature text-only outputs and focus on Visual Question-Answering~(VQA) tasks, limiting their applicability in complex scenarios and failing to sufficiently assess the capability of interleaved image-text processing~\cite{longbench_s3, blau2024gram_ss15, kim2022ocr_ss4, lee2023pix2struct_ss5}.
While some recent efforts have explored multi-page document understanding~\cite{hierarchical_s35, document_s31, slidevqa_s32}, their datasets often lack sufficient complexity in terms of cross-page and cross-element reasoning, whereas the recent MMLongBench-Doc~\cite{mmlongbench_s36} proves excessively difficult, resulting in universally poor performance across models.

To investigate this critical issue and bridge this gap, we introduce a novel and challenging Multimodal Document Summarization Benchmark~\textbf{(M-DocSum-Bench)}, a reference-based generation task, which requires generating interleaved image-text summaries with reference-based images directly from long documents. 
M-DocSum-Bench comprises 500 high-quality paper documents from arXiv primarily across six domains, created in or after April 2024, along with human-preference-aligned multimodal interleaved summaries.
To facilitate evaluation and align with real-world application scenarios, these summaries are divided into four paragraphs, each optionally accompanied by a reference image based on the original image contents.
Each paragraph has at most one accompanying image, and each image appears only once. 
Through these efforts, M-DocSum-Bench offers two key advantages over previous work: 
\textbf{(1) Clarity and intuitiveness:} The reference-based multimodal generation output format intuitively reflects the model's overall understanding of interleaved image-text information. 
As illustrated in Figure~\ref{fig:intro}, the interleaved summarization format enables a clear observation of both image referencing and summarization results, thus providing a foundation for the investigation of model capabilities. 
\textbf{(2) Comprehensive coverage:} The summary encompasses all critical key points from the beginning to the end of the document, including both images and text. 
This rigorously evaluates the comprehensive capabilities of the models such as understanding, reasoning, locating and summarization, enabling in-depth analysis of its performance in handling long-range dependencies, multimodal integration, and global coherence.

To facilitate this Benchmark, we develop an automated framework for constructing interleaved summaries which is subjected to a rigorous development process including cross-validation of human experts. 
Subsequently, we propose a fine-grained evaluation method namely \textbf{M-DocEval}. 
This method centers on three primary dimensions:
\textbf{(1)Textual Content:} M-DocEval evaluates the completeness and accuracy of the textual summary. 
\textbf{(2)Visual Reference:} This dimension assesses the precision and overall effectiveness of image referencing.
\textbf{(3)Instruction Following:} This metric evaluates how well the model follows instructions.
Furthermore, M-DocEval ensures reliability and consistency through a rigorous process, leveraging a powerful LLM such as GPT-4o.

We conduct experiments on M-DocSum-Bench with several leading closed-source models (Gemini Pro~\cite{team2024gemini}, GPT-4o~\cite{openai_gpt4o}, Claude-3.5-Sonnet~\cite{TheC3}, etc.) and powerful open-source models (Qwen2.5-VL-72B~\cite{bai2025qwen2}, Qwen2-VL-72B~\cite{wang2024qwen2}, InternVL2.5-8B~\cite{chen2024expanding}, etc.).
The evaluation results reveal a significant gap between the document summarization capabilities of LVLMs and humans.
We also find that while LVLMs can process individual images and text segments, they struggle to maintain coherence and accurately integrate information within long interleaved contexts, particularly in image understanding, and often exhibiting confusion between similar images and a lack of robustness.

To explore the performance boundary, we further develop a robust summarization baseline, i.e., M-DocSum-7B based Qwen2-vl-7B,
Notably, leveraging a progressive two-stage training approach encompassing instruction-tuning and Direct Preference Optimization~(DPO)~\cite{rafailov2023direct}, along with the diverse instruction and preference data generated by our automated framework, the M-DocSum-7B achieves state-of-the-art performance among a range of closed-source and larger open-source models.
This demonstrates the potential of LVLMs for enhancing interleaved image-text understanding and provides new insights for the document understanding community.

Our contributions are summarized as follows:

\begin{itemize}
\item We introduce a novel and challenging M-DocSum-Bench, addressing a critical gap in the current landscape.
\item We propose an automated evaluation method M-DocEval, providing a realistic and objective assessment.
\item We develop a robust summarization baseline M-DocSum-7B, providing new insights for document understanding.
\end{itemize}

%% file: sec/2_related_work.tex
\section{Related Work}

\subsection{Benchmark for Document Understanding}
Document understanding evaluation has evolved through multiple stages, with studies varying in modality, spatial dimensions, and tasks. Early text-only benchmarks \cite{searchqa_s7, question_s9} like SearchQA, and QuALITY, test long-text processing (1,850-60k tokens) but lack multimodal understanding. ChartQA \cite{chartqa_s15} and others \cite{towards_s28, visualwebbench_s29} 
introduces multimodal inputs, but focuses on isolated components, limiting assessment of comprehensive reasoning.

Most benchmarks have been limited to single-page documents. DocVQA \cite{docvqa_s13}, VisualWebBench \cite{visualwebbench_s29}, OCR-VQA \cite{ocrvqa_s16}, and InfoVQA \cite{infographicvqa_s17} focus on local information extraction (VQA, text extraction). MP-DocVQA \cite{mmlongbench_s36} extends to multi-page, but still avoids cross-page dependencies. SlideVQA \cite{slidevqa_s32} has some cross-page questions (12.9\%) but low information density. DUDE \cite{document_s31} has few cross-page questions (2.1\%) and short contexts (5.3 pages). RULER \cite{ruler_s24} uses distracting information, and FinanceBench \cite{financebench_s34} has long documents and cross-page questions but is domain-specific and relied on costly human evaluation.

Long-context understanding research used synthetic tasks (Needle-in-a-Haystack \cite{needle_s22}, Counting stars \cite{counting_s23}) to test memory, but these lacked real-world complexity. ZeroSCROLLS \cite{zeroscrolls_s1}, LongBench \cite{longbench_s3}, and InfiniteBench \cite{infinite_s4} and others \cite{loogle_s6, marathon_s25} 
offered realistic NLP tasks, but were text-only.

Mmlongbench-doc \cite{mmlongbench_s36} achieved 47.5-page multimodal evaluation, and M-LongDoc \cite{mlongdoc_s14} extended to 210.8 pages with open-ended questions, but lacked refined metrics for locating and summarizing key information. Our proposed DocSum output format addresses this by reflecting the model's overall understanding of interleaved image-text information and comprehensively covers all critical information.

\subsection{Method for Document Understanding}
Document understanding methods have evolved alongside multimodal representation learning and context expansion. Early approaches used cascaded OCR and layout analysis \cite{xu2020layoutlmv2_ss2, huang2022layoutlmv3_ss3}, but this was computationally redundant and not easily optimized end-to-end. OCR-free models \cite{kim2022ocr_ss4,lee2023pix2struct_ss5} addressed this by directly extracting image features and combining them with text embeddings, reducing error accumulation, but still struggled with cross-page visual cues.

To handle long documents, hierarchical encoding \cite{hierarchical_s35,dong2024multi_ss8} was introduced, decomposing documents into page units for separate encoding before LLM processing. However, this could compromise global coherence. Retrieval-augmented generation (RAG) \cite{yu2024visrag_ss13_ss14,blau2024gram_ss15,mmvqa_s20,shi2023replug_ss19} enhanced parsing with external knowledge, and RETRO \cite{borgeaud2022improving_ss18} used chunked cross-attention for long context integration. Recent end-to-end methods \cite{hu2024mplug_ss12} improved long-range dependency modeling with dynamic visual token trimming and large-scale instruction tuning.

In model architecture, advances in long context LVLMs and LLMs have boosted document understanding. Optimizations like sparse attention \cite{liu2023ring_ss21,jaszczur2021sparse_ss22} and state space models \cite{jaszczur2021sparse_ss22,chen2023longlora_ss23,ding2023longnet_ss24} reduced memory overhead, enabling processing of longer contexts. Innovations such as hierarchical representation \cite{gu2021efficiently_ss25}, recursive memory \cite{fu2022hungry_ss26}, and scalable position encoding \cite{beck2025xlstm_ss27} improved capture of spatial-semantic relationships. However, these advancements, while effective for text, face challenges in multi-page multimodal documents, including low cross-modal alignment efficiency and inadequate long-distance visual dependency modeling. We explore LVLM performance boundaries in document summarization using automatically generated preference data, instruction tuning, and DPO.

%% file: sec/3_intersum_bench.tex
\section{M-DocSum-Bench}

\input{tabs/1_statisticians}
\begin{figure}[t]
\centering
\includegraphics[width=0.47\textwidth]{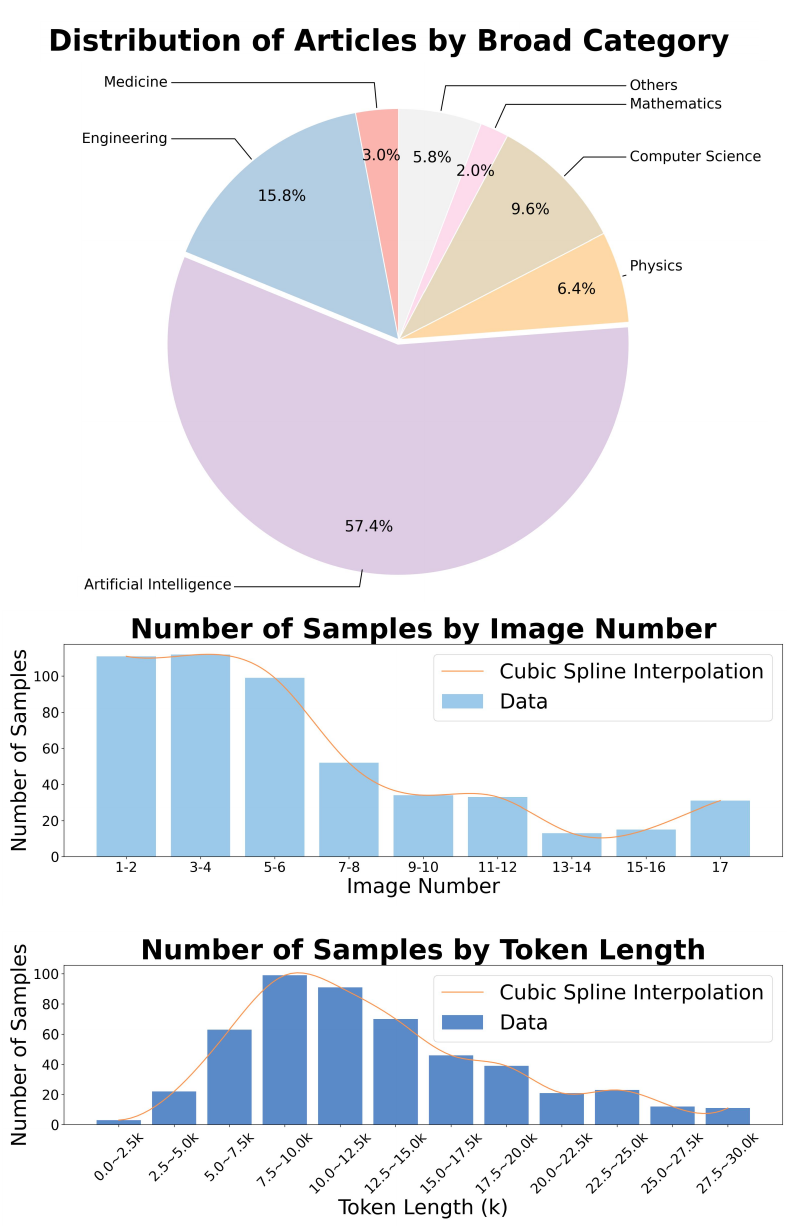}
\caption{The quantitative indicators of M-DocSum-Bench display fundamental information such as token length, image count, and document topics.}
\label{fig:1_statisticians}
\end{figure}

\begin{figure*}[t]
\centering
\includegraphics[width=\textwidth]{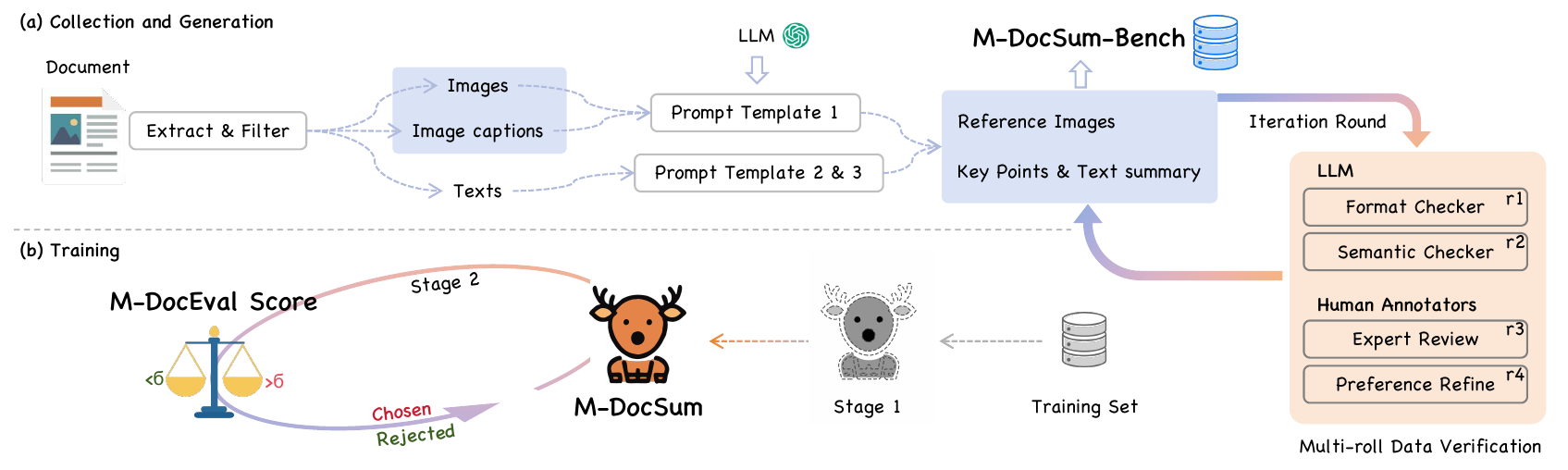}
\caption{The illustration of automated data construction pipeline, multi-roll data verification, and two-stage training.}
\label{fig-main}
\end{figure*}

\subsection{Task Define}
We introduce the M-DocSum task, given the extensive demand for comprehending and summarizing lengthy documents such as scientific articles and technical reports, we have selected scientific articles from arXiv.
The input comprises two modalities, first, the textual information of the scientific article is parsed into complex markdown format, preserving not only the main body of the text but also critical elements like tables and image captions. 
Second, the accompanying images within the article are parsed into image format, retaining their original visual information. 
These two modalities are then fed into the model in an interleaved manner, simulating the natural alternating text-image experience of human readers when engaging with scientific articles. 
Following model processing, a structured output is generated, including: 1) Summary text: a concise summary of the article's core content divided into four fixed paragraphs, covering essential information such as background, main arguments, experimental results or analysis, and conclusions. 
2) Referenced images: the model generates indices for referenced images along with their corresponding captions. The captions show the reasoning when making the selection.
By matching these indices with the original image sequence, we parse the output into the final interleaved summary format. 
This output format not only demands strong text summarization skills from the model but also requires accurate identification and referencing of key images, demonstrating a comprehensive understanding of multimodal information. 
Additionally, the fixed paragraph structure provides clear criteria for evaluating the generation, facilitating both automated and human assessments.

\subsection{Dataset Collection}
\label{sec:3.2}
To support our evaluation benchmark, we process high-quality multimodal documents from the publicly accessible arXiv website~\footnote{\url{https://arxiv.org/}}~\cite{he2024pasa},
primarily focusing on six domains: Artificial Intelligence, and AI-related fields such as Engineering, Computer Science, Physics, Mathematics, and Medicine, given the specialized knowledge required for academic articles. 
To mitigate the risk of data contamination or memorization by existing models, we limit the publication dates to April 2024 or later. 
We filter out documents without images and those exceeding 32k text tokens, ultimately collecting 4.2k high-quality academic articles. 
We carefully select 500 articles for the M-DocSum-Bench for broader benchmark coverage, reserving the remaining 3.7k documents for the training set. 
We have taken steps to ensure the integrity and accuracy of the multimodal data within this benchmark.  
For the evaluation, the images contained within the M-DocSum-Bench are considered authoritative.
As shown in Table~\ref{tab:statis} and Figure~\ref{fig:1_statisticians}, we compile detailed basic information, including document length, number of images, and the proportion of each subject. 
The collection of this data lays the foundation for subsequent exploration of the performance of LVLMs.

\subsection{Interleaved Summary Generation}
\label{sec:3.3}
As shown in Figure~\ref{fig-main}(a), to create a high-quality benchmark, we design a meticulous automated framework for generating summary texts and selecting referenced images. 
Initially, we extract key points for each summary paragraph from the original text, adhering to specific rules: ensuring no repetition of information, maintaining atomicity so each point is an independent, indivisible unit, verifying that each point is traceable back to the original text, and limiting the number of key points per paragraph to no more than 10. 
Once these key points are confirmed, we utilize advanced LLM like GPT-4o to synthesize them into coherent and comprehensive paragraph summaries.

For image referencing, we provide the extracted key points, original images, and image captions to a powerful LVLM.
The model first determines if an image is necessary to aid understanding of the summary paragraph. 
If deemed necessary, it selects the most suitable image based on criteria such as high relevance to the paragraph's content, providing crucial visual support that complements textual information, and ensuring that only one image is referenced per paragraph and each image is used only once. 
Detailed descriptions of our selection and generation processes are available in Appendix Figure~\ref{fig:prompt1},~\ref{fig:prompt2}, and~\ref{fig:prompt3}.

\subsection{Quality Control}
\label{sec:3.4}
As shown in Figure~\ref{fig-main}(right), we introduce Multi-roll Data Verification, a semi-automatic quality control process combining LLMs and human expertise to ensure high benchmark quality. 
Initially, LLMs perform two rounds of validation: Round 1 checks for necessary key fields and at least one image reference in the summary format, while Round 2 ensures semantic integrity by identifying repetitive or invalid content. 
Summaries failing these checks are iteratively regenerated. 
Following LLM validation, domain experts conduct thorough reviews, spending at least 20 minutes on each document to correct any hallucinations or semantic distortions. 
They also refine image references to align with human reading habits. 
Additionally, annotators cross-check each other’s work, with primary reviewer resolving any disagreements to maintain consistency. 
This rigorous process ensures a high-quality, reliable benchmark for advancing multimodal document understanding.

\subsection{M-DocEval}
\label{sec:3.5}
To provide a comprehensive and objective assessment of the performance in generating interleaved summaries, we design a multifaceted set of evaluation metrics. 
These metrics cover textual quality, image referencing appropriateness, and instruction-following capability, ensuring a holistic evaluation of the model abilities.
Specific details regarding text and image reference evaluation can be found in Appendices Figure~\ref{fig:case_en_2} and~\ref{fig:case_en_1}.

\textbf{Textual Content Evaluation.} The text summary evaluation focuses on two primary aspects:
(1) Completeness~(Com) measures the proportion of essential information captured from the original text using a benchmark set of key points, generated and verified by human experts. 
(2) Accuracy~(Acc) assesses the correctness of the summary by comparing each sentence to the original content, rewarding matches and penalizing hallucinations, repetitions, or semantic distortions. 
We compute the Text Score~(TS) as the F1 score of completeness and accuracy:
\begin{equation}
\text{TS} = \frac{2 * \text{Com} * \text{Acc}}{\text{Com} + \text{Acc}}.
\end{equation}
This balanced score ensures a comprehensive assessment of the textual quality.
Specific prompt details can be found in Appendix Figure~\ref{fig:prompt_com} and ~\ref{fig:prompt_acc}.

\textbf{Visual Reference Evaluation.} For image referencing, we evaluate the model's ability to determine image necessity and select appropriate images correctly. 
(1) None Accuracy~(NonAcc) assesses the correct identification of paragraphs needing no image. 
(2) Image Accuracy~(ImgAcc) measures precise image matching for paragraphs requiring images. 
(3) Overall Matching Rate~(OMatch) provides an overall assessment of correct image decisions across all paragraphs. 
(4) Jaccard Similarity~(JacSim) evaluates the similarity between the model's referenced images and the correct set, excluding ``None" cases, offering insight into image selection accuracy regardless of placement. 
The Image Score~(IS) is computed as the weighted average of Overall Matching Rate and Jaccard Similarity:
\begin{equation}
\text{IS} = \frac{\text{OMatch} + \text{JacSim}}{2}.
\end{equation}
This score reflects both the correctness of image selection and the overall appropriateness of image references in the summary, ensuring a robust and objective evaluation framework.

\textbf{Instruction Following Capability.} Given that open-source multimodal models may not always generate content that fully adheres to the given instructions, we introduce an Instruction Following Capability~(IF) metric. 
This metric evaluates how well the model follows instructions by assessing whether it produces appropriate summary texts and image references as specified.

To obtain a final, comprehensive evaluation of each sample, we integrate the TS, IS, and IF using a weighted average:
\begin{equation}
    Total = \alpha * IF + \beta * TS + \gamma * IS,
\end{equation}
where $\alpha, \beta, \gamma$ are $0.1, 0.45, 0.45$, respectively. This weighting scheme ensures that while instruction adherence is important, the quality of the textual and visual content remains the primary focus.

%% file: tabs/1_statisticians.tex
\begin{table}[tbp]
\caption{Overall Statistics of M-DocSum-Bench. This includes the number of documents (Docs), the average number of text tokens (Tokens), and an indicator of whether the documents are multi-page (Multi). The evaluated model capabilities are denoted by U, R, L, S, representing Understanding, Reasoning, Locating, and Summarization, respectively. Data types are classified as Synthetic (generated data), Realistic (purely manually annotated data), and Hybrid (a mix of synthetic and realistic data).}

\centering
\resizebox{0.47\textwidth}{!}{
\begin{tabular}{lcccccccc}
\toprule
Benchmarks & Docs & Tokens & Multi & U & R & L & S & Type \\
\midrule
DocVQA        & -       & 151.5   & $\times$          & $\checkmark$   & $\times$   & $\times$  & $\times$        & synthetic          \\
ChartQA       & -       & 236.9   & $\times$          & $\checkmark$   & $\times$   & $\times$  & $\times$        & synthetic         \\
RULER & any       & any   & $\checkmark$     & $\checkmark$   & $\times$  & $\times$     & $\times$  & synthetic          \\
DUDE & 5019    & 1,831.5 & $\checkmark$     & $\checkmark$   & $\checkmark$   & $\times$  & $\times$   & hybrid          \\
loong  & 1600    & 110,900 & $\checkmark$     & $\checkmark$   & $\checkmark$     & $\times$  & $\times$      & hybrid     \\
MMLongBench  & 135     & 21,214.1 & $\checkmark$    & $\checkmark$      & $\checkmark$      & $\times$     & $\times$ & Realistic       \\
M-Longdoc & 180 & 120,988 & $\checkmark$ & $\checkmark$ & $\checkmark$ & $\times$  & $\times$ & hybrid  \\
DocBench  & 229     & 46,377 & $\checkmark$     & $\checkmark$      & $\checkmark$      & $\times$      & $\times$      & hybrid \\
LongDocURL  & 396 & 43,622.6 & $\checkmark$  & $\checkmark$ & $\checkmark$      & $\checkmark$   & $\times$  & hybrid        \\
\midrule
\textbf{M-DocSum-Bench}            & 500    & 12,912.4 & $\checkmark$   & $\checkmark$     & $\checkmark$   & $\checkmark$   & $\checkmark$  & hybrid        \\
\bottomrule
\end{tabular}
}
\label{tab:statis}
\end{table}

%% file: sec/4_training.tex
\section{Training}
After establishing the InterSum-Bench, as shown in Figure~\ref{fig-main}(b), we initialize M-DocSum-7B with the weights from Qwen2-VL and adopt a two-stage training strategy.

\input{tabs/0_main}

\subsection{Stage 1: Instruction-Tuning}
During this stage, our primary goal is to enhance the accuracy and comprehensiveness of the model in extracting key points for summary generation, while ensuring that the referenced images are appropriate and well-structured.
To achieve this, we utilize the training set generated by the automated framework described in Section~\ref{sec:3.3}.
We train the model for one epoch on the training set using 64 H800 GPUs, configuring all components except the ViT to be trainable. 
For each image, we set the maximum pixels equal to the minimum pixels.
The globe batch size is set to 64, max length is set to 24k, and the learning rate is configured to 1e-5.

\subsection{Stage 2: Direct Preference Optimization}
In this stage, we introduce a novel method for constructing preference data. 
By conducting a second-stage DPO training on this data, we can further enhance the performance and robustness of the model. 

We utilize the model $\pi_\theta$ initialized from the first stage. 
Given an original prompt $x$, $\pi_\theta$ generates an interleaved summary $y$. 
We then modify the original prompt to obtain a degraded output $\widetilde{y}$ as a negative sample. 
The modifications include: (1) Shuffling Image Order: Altering the order of images without providing image indices.
(2) Adding Constraints: Requiring each paragraph to include an image reference, disallowing no selection.
(3) Reducing Information: Providing only the abstract and introduction paragraphs of the original article.
These perturbations increase task complexity while reducing input information. 
The corrupted prompt $\widetilde{x}$ is entered into $\pi_\theta$ to generate a new interleaved summary, forming a preference data pair $\{x=x, y_w=y, y_l=\widetilde{y}\}$, where $y_w, y_l$ are the chosen and rejected sample, respectively.

To prevent the output $\widetilde{y}$ from not being strictly worse than $y$ after being given the prompt $\widetilde{x}$, an effective filtering mechanism is necessary. 
We use M-DocEval proposed in Section~\ref{sec:3.5}, employing Text Score (TS) and Image Score (IS) metrics to screen the preference data.
A preference data pair is considered standard if it satisfies the following formula:
\begin{equation}
    \Delta TS > 0~~\text{and}~~\Delta IS > 0~~and~~\Delta TS + \Delta IS > \delta,
\end{equation}
where the $\Delta TS$ and $\Delta IS$ represent the differences in TS and IS between $y$ and $\widetilde{y}$, and $\delta$ is an adjustable threshold that controls the margin.

We generate preference data using the same articles from the training set as in the first stage. 
The final policy model is optimized using the following loss function:
\begin{equation}
\begin{aligned}
    \mathcal{L}_{\text{DPO}} = -&\mathbb{E}_{(x, y_w, y_l) \sim \mathcal{D}} \\ \Big[ &\log \sigma \Big( \beta \log \frac{\pi_\theta(y_w|x)}{\pi_{\text{ref}}(y_w|x)} - \beta \log \frac{\pi_\theta(y_l|x)}{\pi_{\text{ref}}(y_l|x)} \Big) \Big],
\end{aligned}
\end{equation}
where $\pi_{\text{ref}}(y_w|x)$ denotes the model obtained during the stage 1.

We train M-DocSum-7B on 64 H800 GPUs, loading only one preference data pair per GPU per training step. 
Global batch size is set to 64, the learning rate is set to 1e-6, and $\beta$ is set to 0.2. 
All training processes utilize the DeepSpeed acceleration framework for efficiency.

%% file: tabs/0_main.tex
\begin{table*}[tbp]
\caption{Overall performance comparison of different models on M-DocSum-Bench. The best performance for each metric is indicated in \textbf{bold}, and the best performance among open-source models is \underline{underlined}.}

\centering
\resizebox{0.98\textwidth}{!}{
\begin{tabular}{lcccccccccccc}
\toprule
\multirow{2}{*}{Models} &\multirow{2}{*}{Size} & \multirow{2}{*}{IF} & \multicolumn{3}{c}{Text} & \multicolumn{5}{c}{Image} & \multirow{2}{*}{Total} \\ 
\cmidrule(lr){4-6} \cmidrule(lr){7-11} 
 &  & & Com & Acc & TS & ImgAcc & NonAcc & OMatch & JacSim & IS &  \\ 
\midrule

\multicolumn{11}{l}{\textbf{Close-Source Models}} \\
GPT-4o\cite{openai_gpt4o} & - & 0.998 & 0.531 & 0.696 & 0.602 & \textbf{0.546} & 0.702 & \textbf{0.579} & \textbf{0.696} & \textbf{0.638} & 0.658 \\
Gemini-Pro\cite{team2024gemini} & - & 0.998 & \textbf{0.581} & \textbf{0.835} & \textbf{0.685} & 0.425 & 0.585 & 0.466 & 0.640 & 0.553 & 0.657  \\
Claude-3-5-sonnet\cite{TheC3} & - & \textbf{1.000} & 0.518 & \textbf{0.835} & 0.639 & 0.497 & 0.576 & 0.507 & 0.670 & 0.589 & 0.653 \\
Step-1o-vision-32k\cite{step} & - & 0.882 & 0.410 & 0.814 & 0.545 & 0.407 & 0.784 & 0.504 & 0.647 & 0.576 & 0.593 \\
Moonshot-v1-32k-vision-preview\cite{Moonshot} & - & 0.854 & 0.485 & 0.716 & 0.578 & 0.335 & 0.821 & 0.474 & 0.607 & 0.541 & 0.589 \\
Doubao-vision-pro-32k\cite{doubao} & - & 0.938 & 0.470 & 0.724 & 0.570 & 0.264 & 0.876 & 0.405 & 0.525 & 0.465 & 0.560 \\
\midrule

\multicolumn{11}{l}{\textbf{Open-Source Models}} \\
Qwen2.5-VL\cite{bai2025qwen2} & 72B & 0.980 & 0.501 & 0.639 & 0.562 & 0.320 & 0.712 & 0.419 & 0.615 & 0.517 & 0.583 \\
Qwen2-VL\cite{wang2024qwen2} & 72B & 0.994 & 0.385 & 0.709 & 0.499 & 0.317 & 0.822 & 0.464 & 0.506 & 0.485 & 0.542 \\
Qwen2.5-VL\cite{bai2025qwen2} & 7B & 0.894 & 0.386 & \underline{0.783} & 0.517 & 0.200 & 0.820 & 0.371 & 0.474 & 0.423 & 0.512 \\
Qwen2-VL\cite{wang2024qwen2} & 7B & 0.972 & 0.317 & 0.769 & 0.449 & 0.403 & 0.400 & 0.337 & 0.504 & 0.421 & 0.489 \\
InternVL-2.5\cite{chen2024expanding} & 8B & 0.736 & 0.402 & 0.709 & 0.513 & 0.032 & 0.989 & 0.366 & 0.056 & 0.211 & 0.399 \\
InternVL-2\cite{chen2024internvl} & 8B & 0.700 & 0.408 & 0.753 & 0.529 & 0.011 & \textbf{\underline{0.992}} & 0.354 & 0.026 & 0.190 & 0.394 \\

\midrule
\textbf{M-DocSum} & 7B & \textbf{\underline{1.000}} & \underline{0.553} & 0.770 & \underline{0.644} & \underline{0.542} & 0.716 & \textbf{\underline{0.579}} & \underline{0.693} & \underline{0.636} & \textbf{\underline{0.667}} \\

\bottomrule
\end{tabular}
}
\label{tab:main}
\end{table*}

%% file: sec/5_experiments.tex
\section{Experiments \& Analysis}
In this section, we first evaluate the performance of various models on M-DocSum-Bench using the inference prompt which can be found in the Appendix Figure~\ref{fig:prompt_infer} and compare M-DocSum-7B with several baselines. 
Then, we conduct an ablation study to analyze our key training steps and data.

\subsection{Main Results}
As shown in Table~\ref{tab:main}, M-DocSum-7B outperforms all open-source models in both text and image overall evaluations. 
Notably, in image referencing, M-DocSum-7B even surpasses advanced closed-source models, with its Image Score exceeding Gemini-Pro by 24\% and GPT-4o by 9\%. 
It is also the only model where both Overall Matching and Jaccard Similarity scores exceed 60\%, highlighting its proficiency in image understanding and referencing.
Some notable results include the exceptional performance in text generation completeness and accuracy for Gemini Pro.

Open-source models show that overall scores increase with model parameter size, with the Qwen2.5 series outperforming the Qwen2 series. 
For InternVL, the results indicate that the model is better at generating high-quality text but neglects the understanding and referencing of the images. 
Its None Accuracy score is nearly 1, suggesting that almost all predictions are ``None" which means that no images are selected.

\subsection{Analysis}

\subsubsection{Paragraph-wise Performance Variations}
Figure~\ref{fig:analysis}(a) reveals that models more accurately reference images in earlier paragraphs, with accuracy declining in later sections.
This might be because earlier images are more directly related to the text, while later images could involve more complexity. 
Figure~\ref{fig:analysis}(b) indicates that the task difficulty is well-calibrated with an average score around 0.55, showing distinct difficulty levels across paragraphs. 
Paragraph 4 scores the highest, reflecting strong summarization capabilities, likely due to its structured and key information-rich nature.
Paragraph 2 follows, indicating robust understanding and summarization of specific methods.
Paragraphs 1 and 3 (Background and Experimental Design) score lower, possibly because of their extensive detail and background information, challenging models in extraction and summarization.
The consistent trend observed across all models indicates that our paragraph and difficulty settings are reasonable and our evaluation is stable, leading to highly consistent results.
Detailed results can be found in the Appendix Section~\ref{Paragraph-wise Performance Variations}.

\subsubsection{Influence of Context Length on Performance}
Figures~\ref{fig:analysis}(c) and (d) illustrate a clear trend: as the context length increases, both text and image scores decrease. Notably, the decline in image scores is more precipitous than that of text scores. This disparity suggests that longer contexts disproportionately challenge the models' ability to accurately reference and integrate relevant images, likely stemming from the increased difficulty of localizing the correct visual information within extensive, multi-page documents. Furthermore, a performance gap is observed between open-source and closed-source models. Open-source models exhibit a steeper performance degradation with increasing context length, while closed-source models demonstrate greater robustness. This difference likely reflects the advantages of closed-source models in terms of larger-scale, higher-quality training data, and potentially more sophisticated training methodologies.
Detailed results can be found in the Appendix Section~\ref{Influence of Context Length on Performance}.

\subsubsection{Impact of Image Quantity on Performance}
\begin{figure*}[t]
\centering
\includegraphics[width=1.0\textwidth]{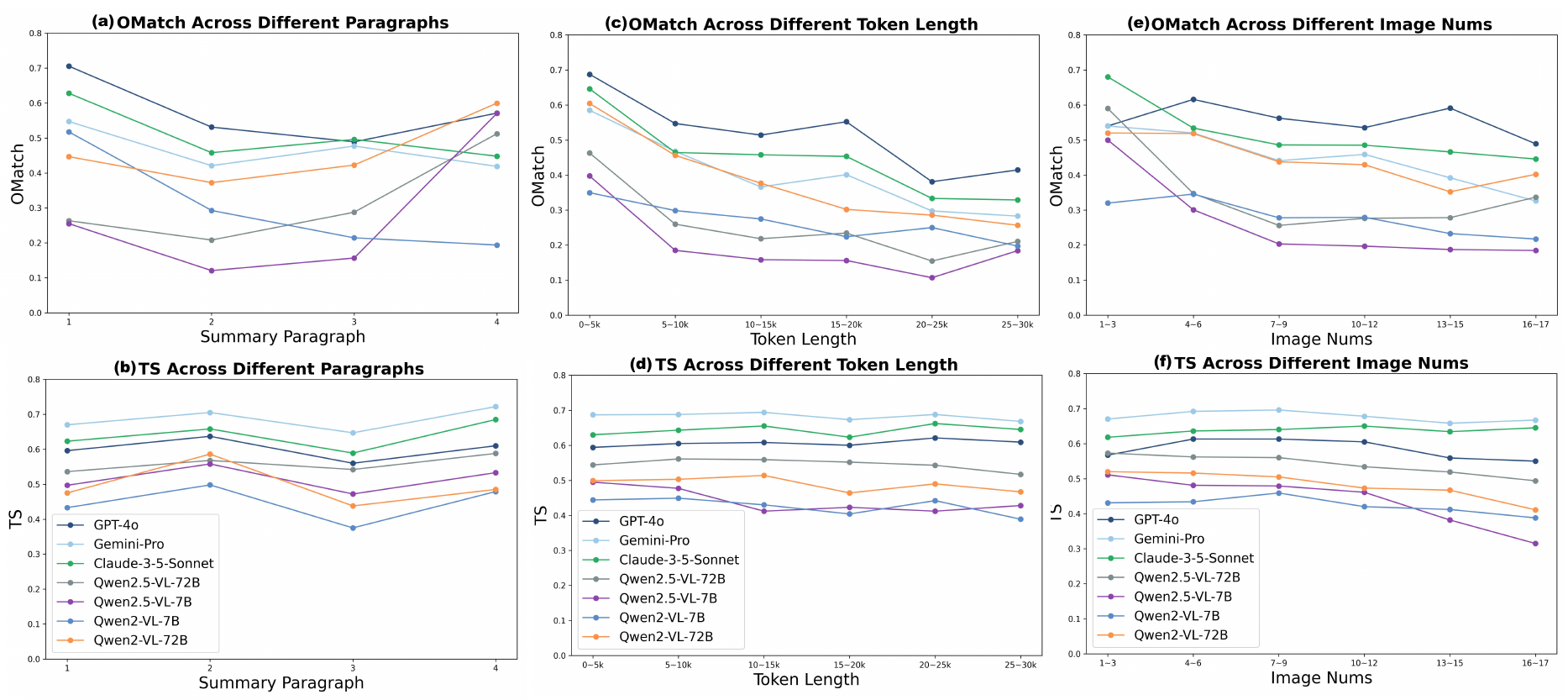}
\caption{Quantitative analysis of the performance trends of different models as data characteristics vary, including the different paragraphs in the interleaved summarization, the token length of the original text, and the number of input images.}
\label{fig:analysis}
\end{figure*}

Figure~\ref{fig:analysis}(e) reveals a significant decrease in image referencing scores across all models as the number of images per document increases. This highlights the challenge models face in selecting pertinent images for summarization in image-rich contexts. The highest scores are observed with documents containing only 1-3 images, suggesting that correct image selection is facilitated in these less visually complex scenarios.
Figure~\ref{fig:analysis}(f) illustrates that leading closed-source models sustain stable text scores irrespective of the image count. This consistency in text generation likely reflects their advanced architectural designs and optimization strategies. Conversely, open-source models, along with certain closed-source models, show a deterioration in text quality as the number of images grows, suggesting limitations in their multimodal integration capabilities, potentially attributable to constraints in model size and training data.
Detailed results can be found in the Appendix Section~\ref{Impact of Image Quantity on Performance}.

\subsubsection{Image Reference Bias}
\input{tabs/2_bias}
Notably, as shown in Table~\ref{tab:bias}, when assuming all output referenced images are ``None", the OMatch score is 0.325, indicating that only 32.5\% of paragraphs actually do not require images. Considering all image metrics together, when both ImgAcc and JacSim are closer to 0 and NonAcc is closer to 1, it indicates that the model has a greater tendency not to select images. This extreme phenomenon explains the lower image-related scores for LVLMs such as InternVL-2 and InternVL-2.5.

\begin{figure}[t]
\centering
\includegraphics[width=0.49\textwidth]{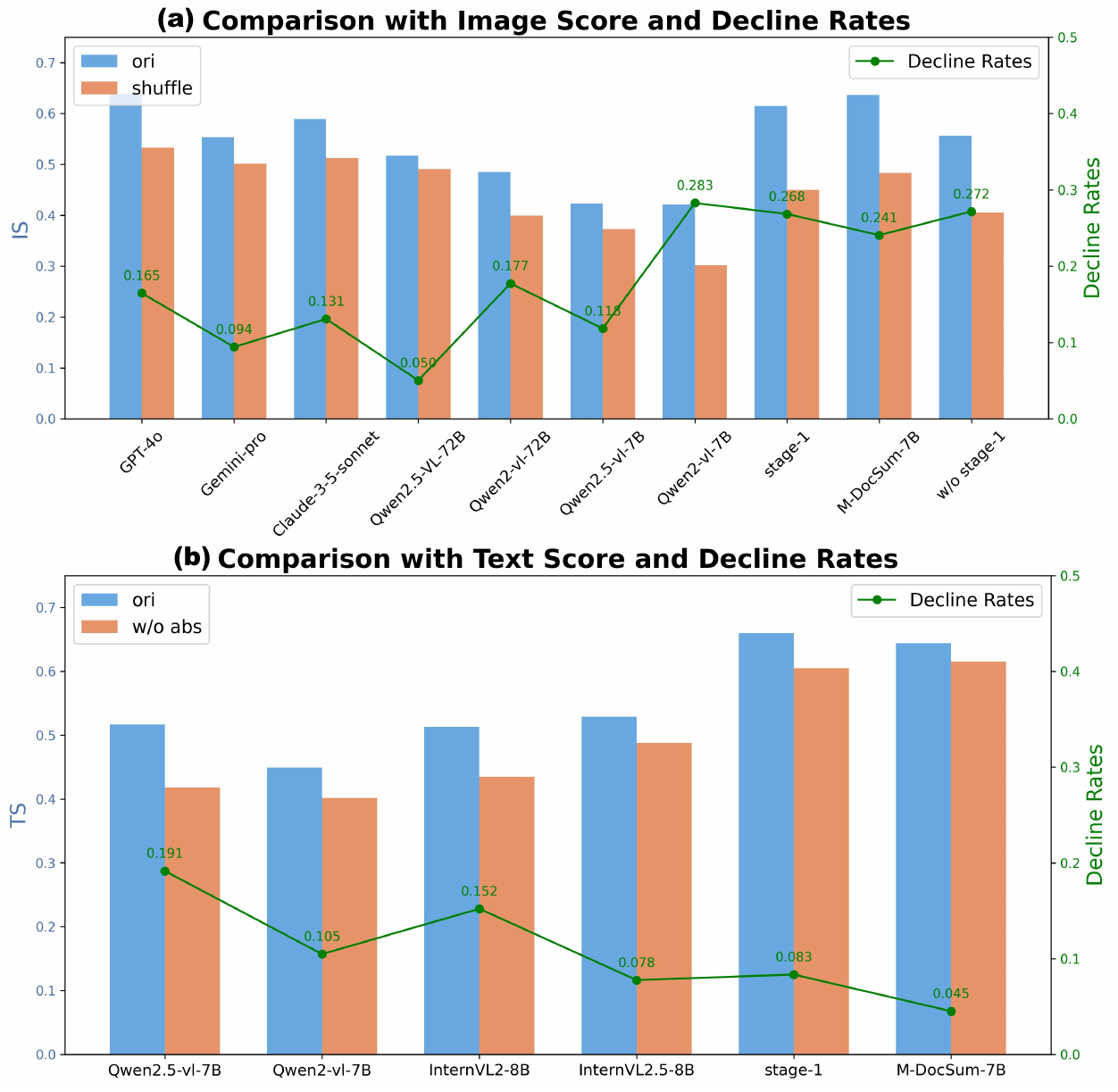}
\caption{Blue bars represent original image scores, orange bars represent scores after modification, and the green line indicates the decline rate.}
\label{fig:ablation}
\end{figure}

\subsection{Ablation Study}
In our ablation study, we test the robustness of LVLMs by shuffling the order of images and removing the original abstract, observing the performance under this out-of-distribution (OOD) condition.
A fundamental hypothesis is that models should primarily rely on the semantic association between images and text rather than the positional arrangement of images. 
If a model accurately captures these semantic relationships, its performance should remain stable even when the image order is changed; conversely, it indicates the model may merely have learned specific patterns.
As shown in Figure~\ref{fig:ablation}(a), overall, closed-source models or larger models exhibit lower sensitivity to image position, with performance decline rates all below 17.7\%. 
In contrast, smaller open-source models, such as Qwen2-VL-7B, demonstrate the poorest robustness. After the first stage of training, while overall performance improves, decline rates are reduced, indicating an enhanced understanding of interleaved image-text information, but not eliminating the dependence on image order. 
With the second stage of training, M-DocSum-7B exhibits even greater robustness. 
Furthermore, experimental results demonstrate that effective first-stage training is essential, if second-stage training is conducted directly, the model's decline rates actually increase.

Another question is whether the model completes the summarization by directly copying the original abstract. 
To investigate this, we conduct experiments on models of the same size with strong textual capabilities.
As shown in Figure~\ref{fig:ablation}(b), the results show that as we progressively train the model, the text score gradually increases while the decline rates decrease.
This effectively demonstrates the effectiveness of the training and indicates the continuous improvement of the robustness.
Furthermore, compared to the disturbance caused by image changes, the absence of the abstract has a weaker impact on the summarization ability, which indirectly confirms the model's low dependence on the original abstract.
Detailed results can be found in the Appendix Section~\ref{Ablation}.

\subsection{Case Study}
We provide two detailed case studies in the Appendix, Figure~\ref{fig:case_en_2} and~\ref{fig:case_en_1}.
Notably, the order of images presented in the case studies may differ from the order in the abstract. 
This discrepancy underscores a common challenge in current vision-language tasks: models often struggle to accurately capture fine-grained correspondences between images and text. 
Furthermore, the case studies clearly illustrate that existing models are prone to confusion when dealing with images that are semantically similar but possess distinct visual details, leading to inaccurate reasoning results. 

%% file: tabs/2_bias.tex
\begin{table}[tbp]
\caption{All image reference results are ``None".}

\centering
\resizebox{0.45\textwidth}{!}{
\begin{tabular}{lccccc}
\toprule
& ImgAcc & NonAcc & OMatch & JacSim & IS  \\
\midrule
``None" &  0 & 1.000 & 0.325 & 0 & 0 \\
\bottomrule

\end{tabular}
}
\label{tab:bias}
\end{table}

%% file: sec/6_conclusion.tex
\section{Conclusion}
This paper introduces the M-DocSum-Bench, a novel and challenging Multimodal Document Summarization Benchmark designed to evaluate the ability of LVLMs to understand interleaved image-text documents.
The benchmark requires generating reference-based, interleaved image-text summaries from 500 high-quality arXiv papers, providing a more comprehensive and intuitive assessment than existing benchmarks.
A fine-grained evaluation method M-DocEval is proposed to assess textual content, visual reference accuracy, and instruction following ability, leveraging a powerful LLM for reliable and consistent results. 
Experiments with leading LVLMs reveal a significant performance gap compared to humans, highlighting challenges in maintaining coherence and integrating information in long interleaved contexts.
Finally, a strong summarization baseline, M-DocSum-7B, is developed using a progressive two-stage training approach (instruction-tuning and DPO). This model achieves state-of-the-art performance among various models, demonstrating the potential of LVLMs for enhanced interleaved image-text understanding.

\section*{Acknowledgements}
We thank Yanan Wei, Yingmin Li, Aihu Zhang, Dingxin Wang for engineering support for this work.

%% file: sec/7_supp.tex
\clearpage
\setcounter{page}{1}

\begin{figure*}[t]
\centering
\includegraphics[width=0.8\textwidth]{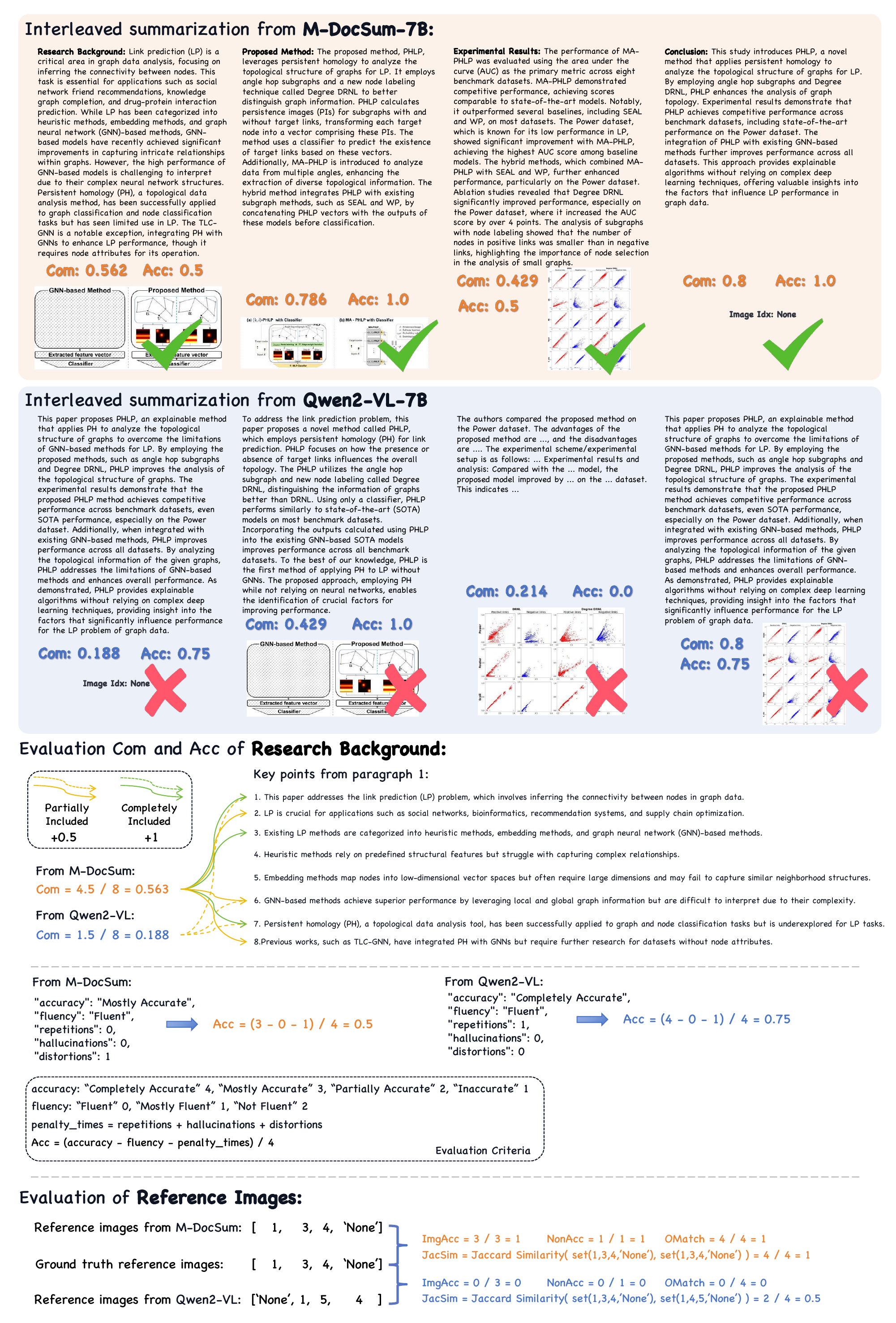}
\caption{Presentation 2 of Case Study and M-DocEval Evaluation Methodology}
\label{fig:case_en_1}
\end{figure*}

\twocolumn[{
\renewcommand\twocolumn[1][]{#1}
\section{Case Study and Evaluation}
We present cases generated by M-DocSum-7B and Qwen2-VL-7B in Figure~\ref{fig:case_en_2} and Figure~\ref{fig:case_en_1}, and elaborate on the calculation methods of the M-DocEval metrics.

\begin{center}
    \centering
    \captionsetup{type=figure}
    \includegraphics[width=0.77\textwidth]{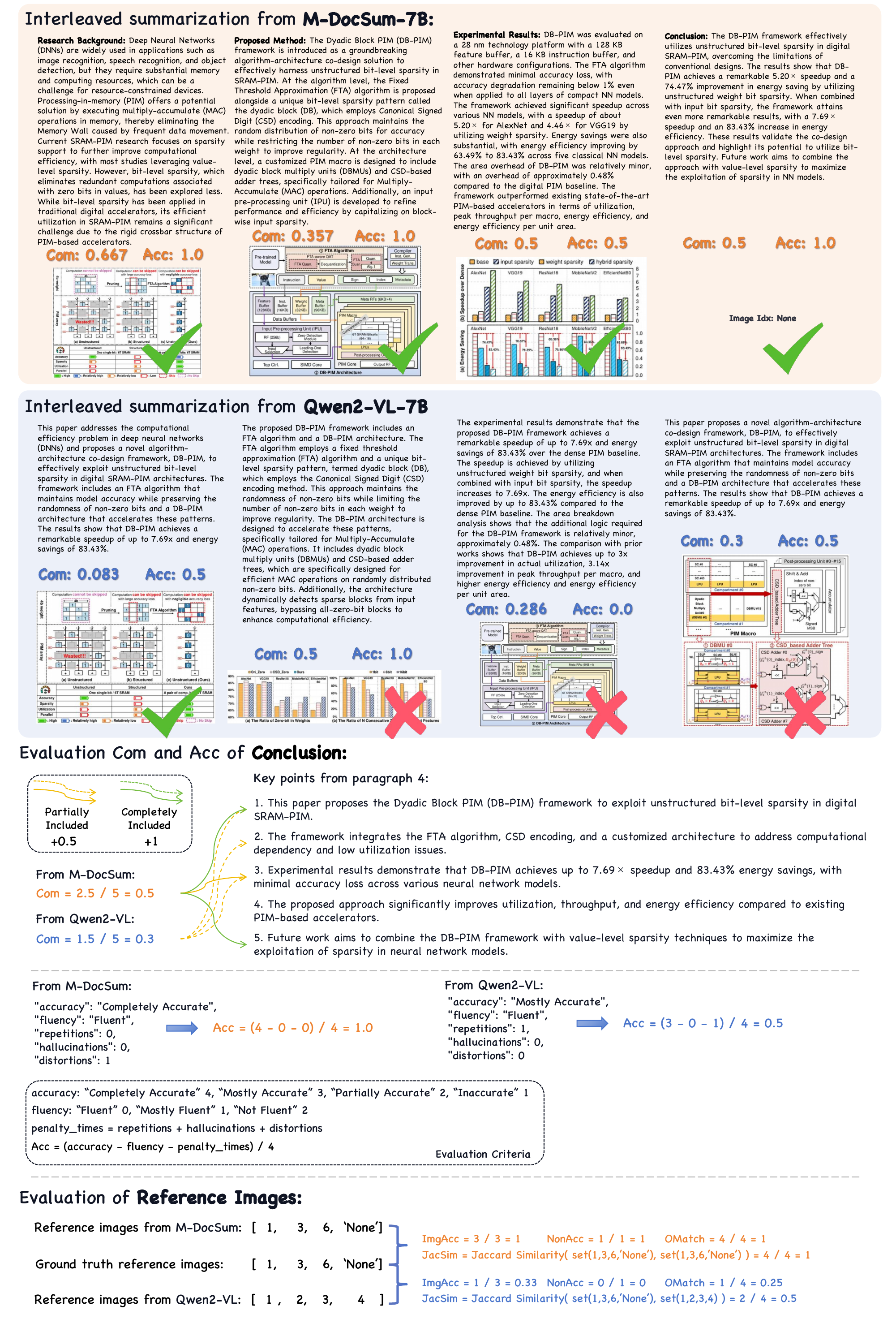}
    \captionof{figure}{Presentation 1 of Case Study and M-DocEval Evaluation Methodology.}
    \label{fig:case_en_2}
\end{center}
}]



\begin{figure*}[t]
\centering
\includegraphics[width=0.9\textwidth]{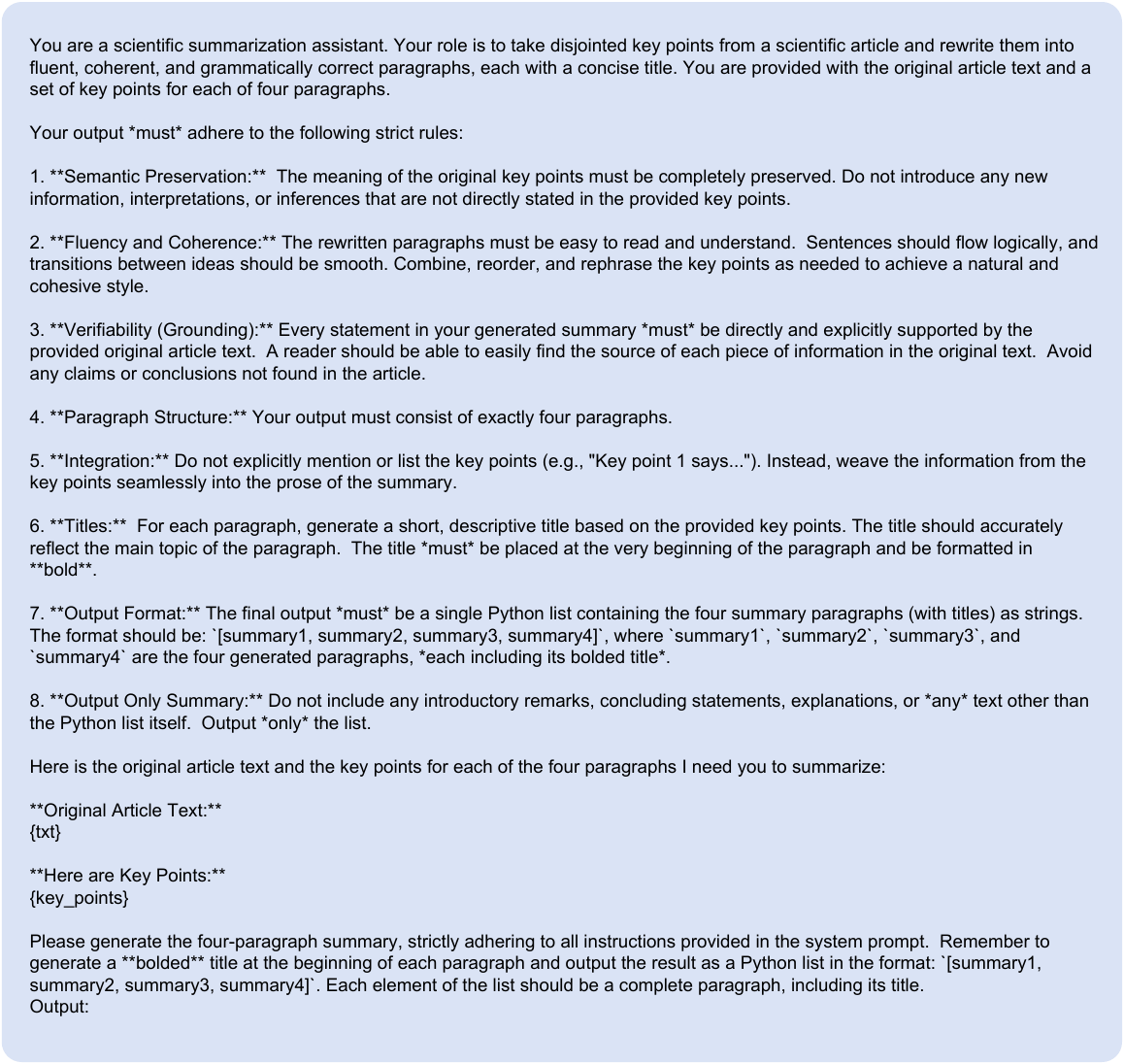}
\caption{Prompt template for summary generation.}
\label{fig:prompt2}
\end{figure*}

\begin{figure*}[t]
\centering
\includegraphics[width=0.9\textwidth]{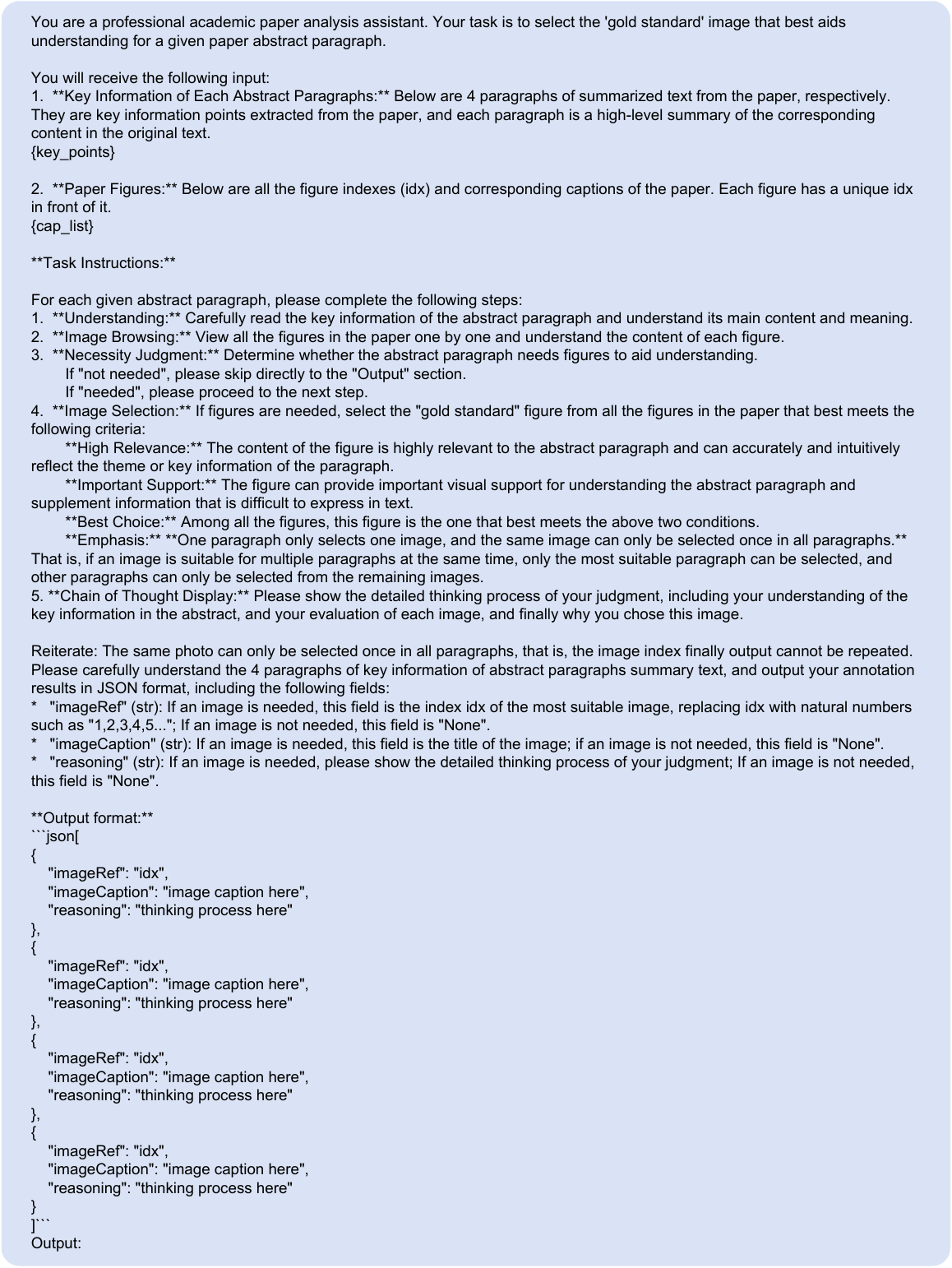}
\caption{Prompt template for reference image extraction.}
\label{fig:prompt3}
\end{figure*}

\begin{figure*}[t]
\centering
\includegraphics[width=0.9\textwidth]{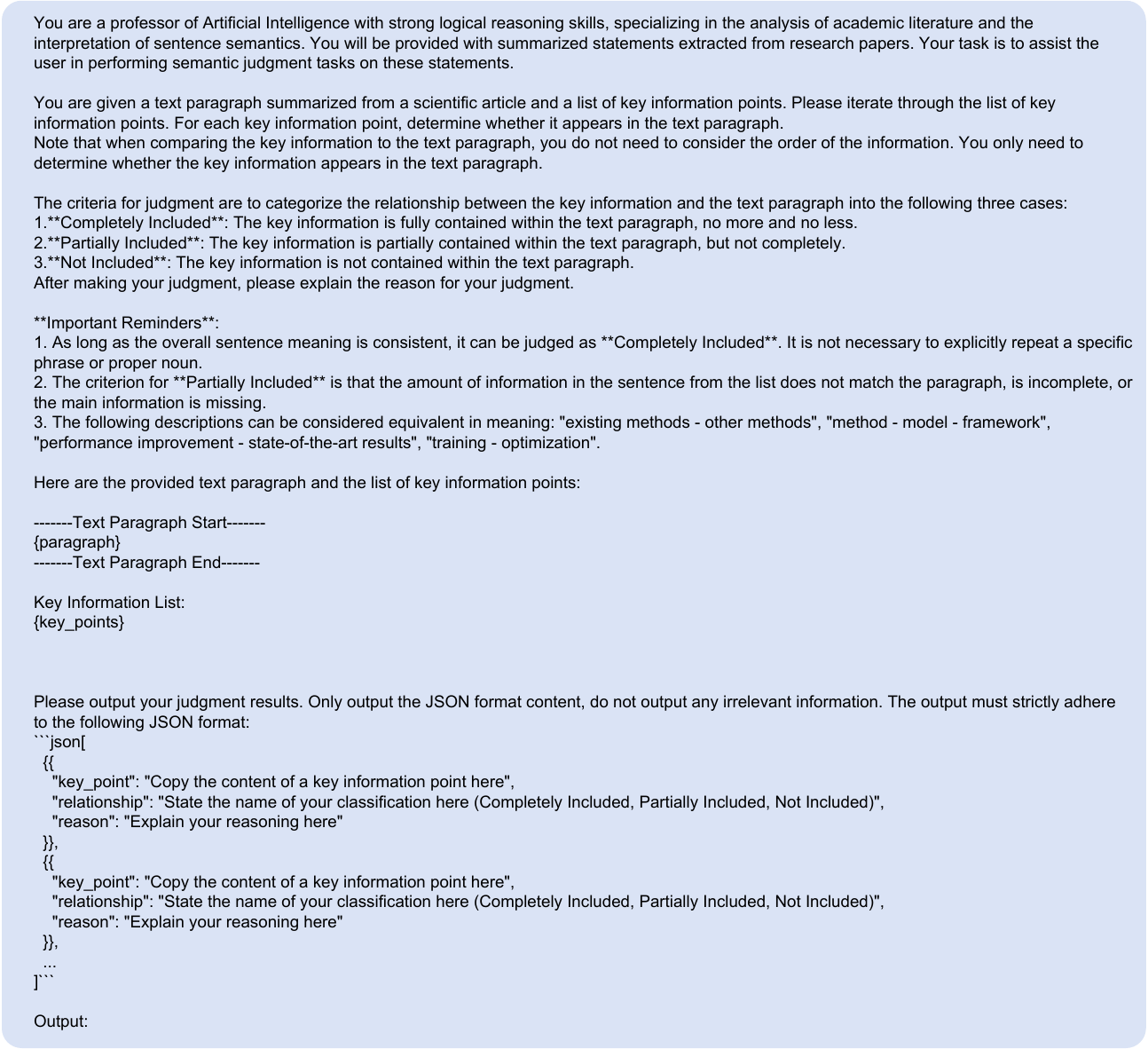}
\caption{Prompt template for evaluating text Completeness (Com).}
\label{fig:prompt_com}
\end{figure*}

\begin{figure*}[t]
\centering
\includegraphics[width=0.8\textwidth]{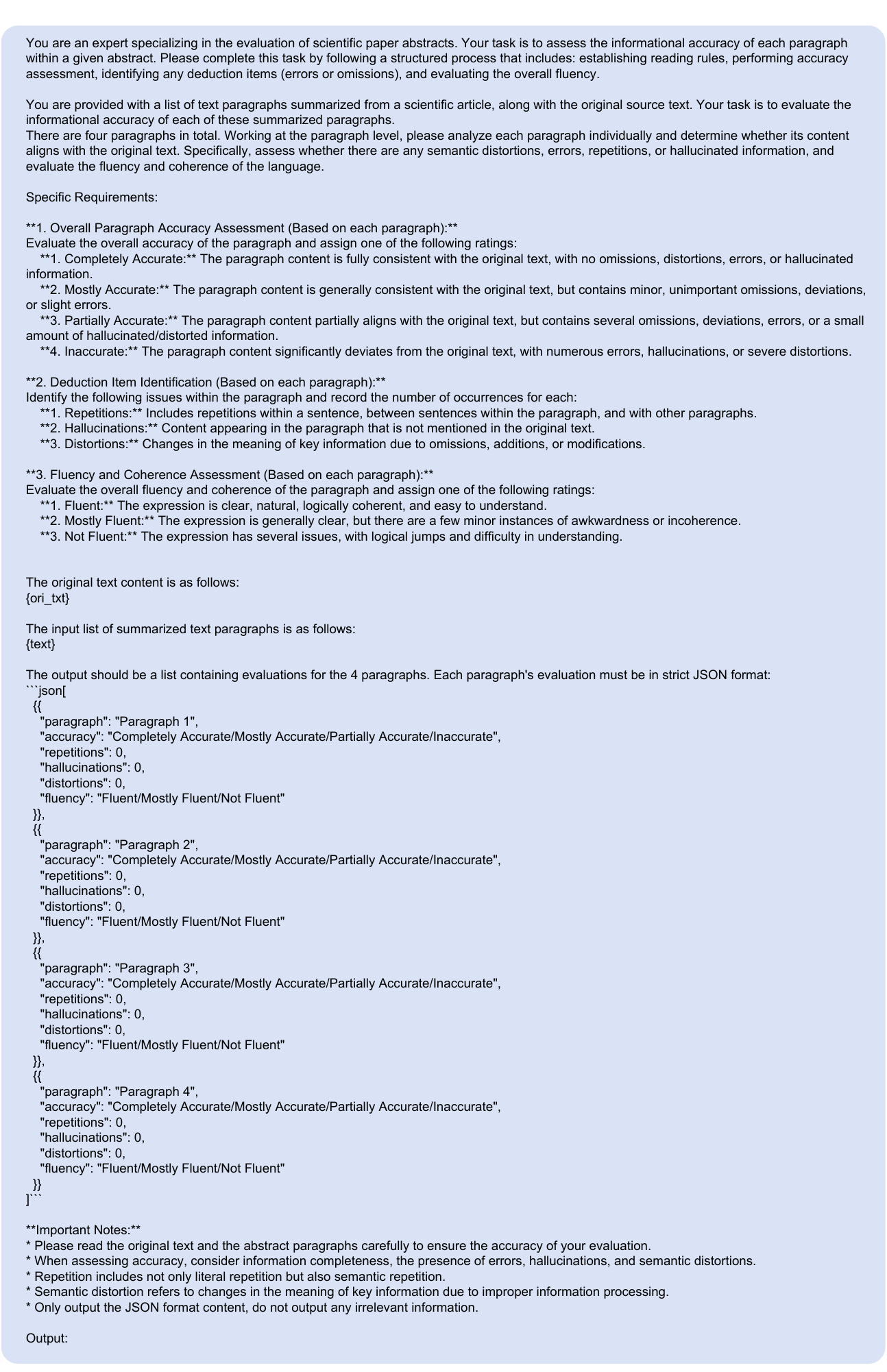}
\caption{Prompt template for evaluating text Accuracy (Acc).}
\label{fig:prompt_acc}
\end{figure*}

\begin{figure*}[t]
\centering
\includegraphics[width=0.8\textwidth]{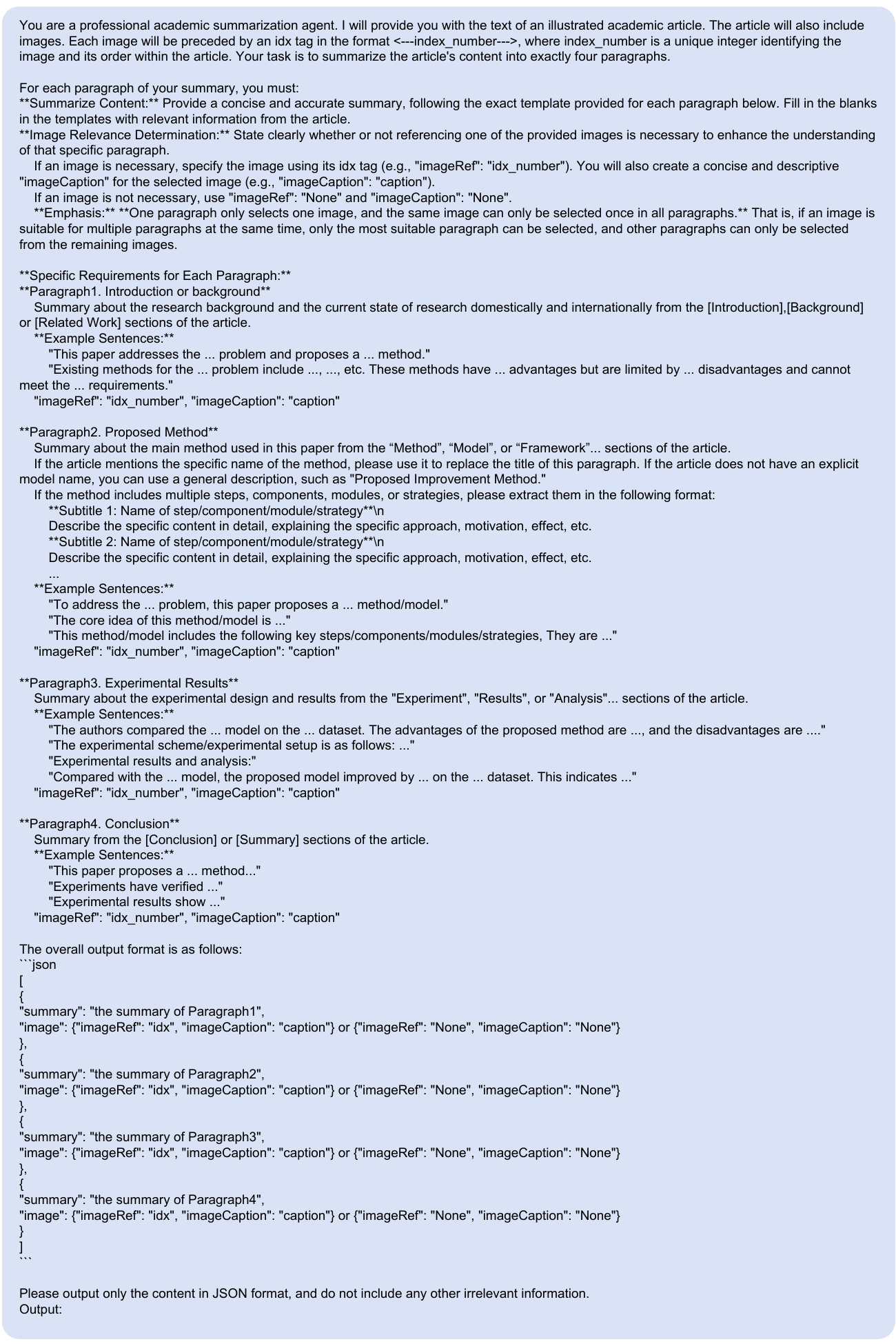}
\caption{Prompt template for inference.}
\label{fig:prompt_infer}
\end{figure*}

\twocolumn[{
\renewcommand\twocolumn[1][]{#1}
\section{Prompt Template}
We list the prompt templates used in all processes of the paper, including the automatic data construction, inference, and evaluation stages, in Figures~\ref{fig:prompt1},~\ref{fig:prompt2},~\ref{fig:prompt3},~\ref{fig:prompt_com},~\ref{fig:prompt_acc} and~\ref{fig:prompt_infer}.

\begin{center}
    \centering
    \captionsetup{type=figure}
    \includegraphics[width=0.83\textwidth]{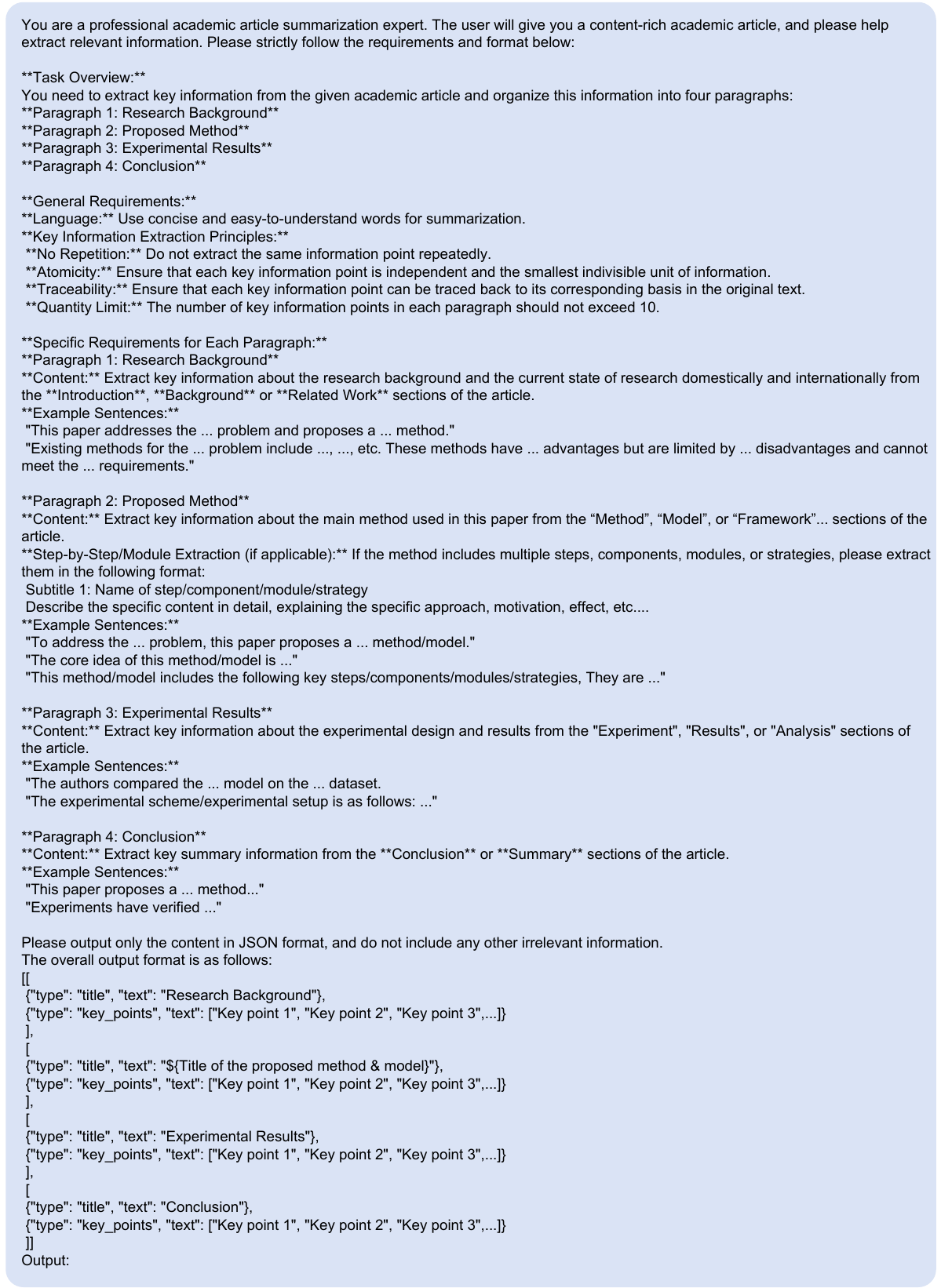}
    \captionof{figure}{Prompt template for key point extraction.}
    \label{fig:prompt1}
\end{center}
}]



\twocolumn[
\section{Analysis}
\subsection{Paragraph-wise Performance Variations}
\label{Paragraph-wise Performance Variations}
The detailed data in the Table~\ref{tab:para} shows that Gemini Pro, while performing best in summarizing text, maintains only an average level of performance in image referencing. For the Qwen-VL series, upgrading from version 2 to 2.5 results in an overall improvement in text performance, but a varying degree of decline in image referencing is observed.

\subsection{Influence of Context Length on Performance}
\label{Influence of Context Length on Performance}
The detailed data in Table~\ref{tab:token} shows that GPT-4o has a commanding lead in image-related aspects, and Gemini Pro has a commanding lead in text-related aspects, and these leading trends do not diminish as the text length increases. Notably, regarding image referencing capability, the scores of all models decrease by approximately half when the text length increases to 30k compared to the 0-5k range. This is a very pronounced trend.

\subsection{Impact of Image Quantity on Performance}
\label{Impact of Image Quantity on Performance}
From Table~\ref{tab:image}, it is evident that there is a significant difference in the decline of image referencing performance between open-source and closed-source models as the number of images increases. Among the closed-source models, only GPT-4o exhibits a relatively small decline of 9.4\%, while other closed-source models, Claude-3.5 and Gemini-Pro, show declines of 34.4\% and 39.6\%, respectively. Surprisingly, the open-source models Qwen2.5-VL-7B and Qwen2.5-VL-72B experience astonishing declines of 56.6\% and 62.6\%, respectively.

\section{Ablation}
\label{Ablation}
Table~\ref{tab:shuffle} shows that while the robustness of M-DocSum-7B is stronger than that of the base model (+37.5\%) and the model after the first stage of training (+6.8\%), there is still a slight gap compared to powerful closed-source models and larger open-source models. Specifically, the performance after shuffling is lower than that of GPT-4o (-9.4\%), Claude-3.5 (-5.7\%), Gemini Pro (-3.6\%), and Qwen2.5-VL-72B (-1.6\%), indicating that this is an issue worth further investigation in the future.

When we remove the abstract from the original text, the overall performance decline across models is not significant. The largest and smallest declines are observed in Qwen2.5-VL-7B (19.1\%) and M-DocSum-7B (4.5\%), respectively. Notably, the normal score of the model after the first stage of training is higher than that of M-DocSum-7B by 2.4\%, but its performance after removing the abstract is lower by 1.6\%. This provides strong evidence that the second stage of training enhances the model's robustness.
]

\input{tabs/A_1en_combined_para}
\input{tabs/A_2en_combined_token_length}
\input{tabs/A_3en_combined_image_nums}



\input{tabs/A6-A4-A5}

%% file: tabs/A_1en_combined_para.tex
\begin{table*}[t]
\caption{TS and OMatch Across Different Models and Paragraphs}
\centering
\resizebox{0.85\textwidth}{!}{
\begin{tabular}{l|cc|cc|cc|cc}
\toprule
\multirow{2}{*}{Model} & \multicolumn{2}{c|}{para 1} & \multicolumn{2}{c|}{para 2} & \multicolumn{2}{c|}{para 3} & \multicolumn{2}{c}{para 4} \\
\cmidrule(lr){2-3} \cmidrule(lr){4-5} \cmidrule(lr){6-7} \cmidrule(lr){8-9}
 & TS & OMatch & TS & OMatch & TS & OMatch & TS & OMatch \\
\midrule
GPT-4o & 0.596 & 0.705 & 0.637 & 0.531 & 0.560 & 0.489 & 0.610 & 0.571 \\
Gemini-Pro & 0.670 & 0.547 & 0.705 & 0.421 & 0.647 & 0.477 & 0.722 & 0.419 \\
Claude-3-5-Sonnet & 0.623 & 0.628 & 0.658 & 0.458 & 0.589 & 0.496 & 0.685 & 0.448 \\
Qwen2.5-VL-72B & 0.536 & 0.263 & 0.568 & 0.208 & 0.542 & 0.288 & 0.588 & 0.512 \\
Qwen2-VL-72B & 0.475 & 0.447 & 0.586 & 0.372 & 0.438 & 0.423 & 0.485 & 0.600 \\
Qwen2.5-VL-7B & 0.497 & 0.255 & 0.558 & 0.121 & 0.472 & 0.157 & 0.533 & 0.570 \\
Qwen2-VL-7B & 0.433 & 0.518 & 0.498 & 0.293 & 0.375 & 0.214 & 0.479 & 0.194 \\

\bottomrule
\end{tabular}
}
\label{tab:para}
\end{table*}

%% file: tabs/A_2en_combined_token_length.tex
\begin{table*}[t]
\caption{TS and OMatch Across Different Models and Token Length}
\centering
\resizebox{0.95\textwidth}{!}{
\begin{tabular}{l|cc|cc|cc|cc|cc|cc}
\toprule
\multirow{2}{*}{Model} & \multicolumn{2}{c|}{0\textasciitilde5k} & \multicolumn{2}{c|}{5\textasciitilde10k} & \multicolumn{2}{c|}{10\textasciitilde15k} & \multicolumn{2}{c|}{15\textasciitilde20k} & \multicolumn{2}{c|}{20\textasciitilde25k} & \multicolumn{2}{c}{25\textasciitilde30k} \\
\cmidrule(lr){2-3} \cmidrule(lr){4-5} \cmidrule(lr){6-7} \cmidrule(lr){8-9} \cmidrule(lr){10-11} \cmidrule(lr){12-13}
 & TS & OMatch & TS & OMatch & TS & OMatch & TS & OMatch & TS & OMatch & TS & OMatch \\
\midrule
GPT-4o & 0.594 & 0.688 & 0.605 & 0.547 & 0.608 & 0.514 & 0.600 & 0.552 & 0.621 & 0.381 & 0.609 & 0.414 \\
Gemini-Pro & 0.687 & 0.585 & 0.688 & 0.468 & 0.694 & 0.366 & 0.673 & 0.401 & 0.688 & 0.298 & 0.668 & 0.283 \\
Claude-3-5-Sonnet & 0.630 & 0.646 & 0.643 & 0.464 & 0.655 & 0.458 & 0.623 & 0.453 & 0.662 & 0.333 & 0.645 & 0.329 \\
Qwen2.5-VL-72B & 0.544 & 0.463 & 0.561 & 0.260 & 0.559 & 0.218 & 0.552 & 0.234 & 0.543 & 0.155 & 0.517 & 0.211 \\
Qwen2-VL-72B & 0.499 & 0.604 & 0.503 & 0.456 & 0.514 & 0.377 & 0.464 & 0.302 & 0.490 & 0.286 & 0.467 & 0.257 \\
Qwen2.5-VL-7B & 0.495 & 0.397 & 0.477 & 0.185 & 0.412 & 0.158 & 0.423 & 0.156 & 0.412 & 0.107 & 0.428 & 0.184 \\
Qwen2-VL-7B & 0.444 & 0.350 & 0.449 & 0.299 & 0.430 & 0.275 & 0.404 & 0.224 & 0.442 & 0.250 & 0.389 & 0.197 \\

\bottomrule
\end{tabular}
}
\label{tab:token}
\end{table*}

%% file: tabs/A_3en_combined_image_nums.tex
\begin{table*}[h]
\caption{TS and OMatch Across Different Models and Image Numbers}
\centering
\resizebox{0.95\textwidth}{!}{
\begin{tabular}{l|cc|cc|cc|cc|cc|cc}
\toprule
\multirow{2}{*}{Model} & \multicolumn{2}{c|}{1\textasciitilde3} & \multicolumn{2}{c|}{4\textasciitilde6} & \multicolumn{2}{c|}{7\textasciitilde9} & \multicolumn{2}{c|}{10\textasciitilde12} & \multicolumn{2}{c|}{13\textasciitilde15} & \multicolumn{2}{c}{16\textasciitilde17} \\
\cmidrule(lr){2-3} \cmidrule(lr){4-5} \cmidrule(lr){6-7} \cmidrule(lr){8-9} \cmidrule(lr){10-11} \cmidrule(lr){12-13}
 & TS & OMatch & TS & OMatch & TS & OMatch & TS & OMatch & TS & OMatch & TS & OMatch \\
\midrule
GPT-4o & 0.567 & 0.540 & 0.613 & 0.616 & 0.613 & 0.562 & 0.605 & 0.535 & 0.559 & 0.591 & 0.550 & 0.489 \\
Gemini-Pro & 0.670 & 0.540 & 0.692 & 0.520 & 0.696 & 0.441 & 0.678 & 0.459 & 0.658 & 0.392 & 0.667 & 0.326 \\
Claude-3-5-Sonnet & 0.618 & 0.680 & 0.636 & 0.534 & 0.640 & 0.486 & 0.650 & 0.485 & 0.634 & 0.466 & 0.645 & 0.446 \\
Qwen2.5-VL-72B & 0.573 & 0.590 & 0.562 & 0.347 & 0.560 & 0.256 & 0.534 & 0.276 & 0.519 & 0.278 & 0.494 & 0.337 \\
Qwen2-VL-72B & 0.520 & 0.520 & 0.516 & 0.519 & 0.505 & 0.438 & 0.473 & 0.429 & 0.467 & 0.352 & 0.411 & 0.402 \\
Qwen2.5-VL-7B & 0.511 & 0.500 & 0.481 & 0.301 & 0.479 & 0.203 & 0.461 & 0.197 & 0.382 & 0.188 & 0.315 & 0.185 \\
Qwen2-VL-7B & 0.431 & 0.320 & 0.434 & 0.346 & 0.459 & 0.278 & 0.420 & 0.279 & 0.412 & 0.233 & 0.388 & 0.217 \\

\bottomrule
\end{tabular}
}
\label{tab:image}
\end{table*}

%% file: tabs/A6-A4-A5.tex
\begin{table*}[t] 
\centering
\begin{minipage}{0.4\textwidth}
  \centering
  \caption{Image Score and Decline Rates (Shuffled)} 
  \resizebox{\textwidth}{!}{
    \begin{tabular}{lcc}
    \toprule
    Models                      & Original IS & Shuffled IS \\
    \midrule
    GPT-4o                   & 0.638          & 0.533           \\
    Gemini-pro                 & 0.553          & 0.501           \\
    Claude-3-5-sonnet & 0.589          & 0.512           \\
    Qwen2.5-VL-72B             & 0.517          & 0.491           \\
    Qwen2-vl-72B               & 0.485          & 0.399           \\
    Qwen2.5-vl-7B               & 0.423          & 0.373           \\
    Qwen2-vl-7B                & 0.421          & 0.302           \\
    Stage-1       & 0.615          & 0.450           \\
    M-DocSum-7B      & 0.636          & 0.483           \\
    M-DocSum-7B w/o stage-1                 & 0.556          & 0.405         \\
    \bottomrule
    \end{tabular}
  }
  \label{tab:shuffle}
\end{minipage}\hspace{0.05\textwidth}
\begin{minipage}{0.4\textwidth}
  \centering
  \caption{Text Score and Decline Rates (W/o Abstract)} 
  \resizebox{\textwidth}{!}{
    \begin{tabular}{lcc}
    \toprule
    Models                     & Original TS & W/o abstract TS \\
    \midrule
    Qwen2.5-VL-7B                       & 0.517	&0.418           \\
    Qwen2-VL-7B         &  0.449&	0.402           \\
    InternVL2-8B & 0.513&	0.435          \\
    InternVL2.5-8B             & 0.529&	0.488           \\
    Stage-1                & 0.660&	0.605          \\
    M-DocSum-7B               & 0.644&	0.615         \\
    \bottomrule
    \end{tabular}
  }
  \label{tab:abs}
\end{minipage}
\end{table*}